\documentclass[letterpaper]{article}
\usepackage{proceed2e}
\usepackage[margin=1in]{geometry}
\usepackage{hyperref}
\usepackage[english]{babel}
\usepackage[utf8x]{inputenc}
\usepackage[T1]{fontenc}
\usepackage{amsmath,amssymb}
\usepackage{mathtools}
\usepackage{tabularx}
\usepackage[usenames,dvipsnames]{xcolor}
\usepackage{booktabs}
\usepackage{pgfplotstable}
\usepackage{filecontents}
\usepackage{csvsimple}
\usepackage{enumerate}
\usepackage[linesnumbered,ruled]{algorithm2e}
\mathtoolsset{showonlyrefs}

\newcommand{\dataset}{{\cal D}}

\newcommand{\EE}[2]{\mathbb{E}_{#1}\left[ #2 \right]}
\newcommand{\set}[1]{\left\lbrace #1 \right\rbrace}
\newcommand{\bs}[1]{\boldsymbol{#1}}
\newcommand{\bx}{\boldsymbol{x}}
\newcommand{\bz}{\boldsymbol{z}}

\newcommand{\bmu}{\boldsymbol{\mu}}
\newcommand{\btheta}{\boldsymbol{\theta}}
\newcommand{\bomega}{\boldsymbol{\omega}}
\newcommand{\bg}{\boldsymbol{g}}

\newcommand{\gs}{\boldsymbol{g}_s}

\newcommand{\PG}{p_{\scriptscriptstyle \mathrm{PG}}}

\newcommand{\R}{\mathbb{R}}

\newcommand{\X}{\mathcal{X}}
\newcommand{\Z}{\mathcal{Z}}
\newcommand{\LB}{\mathcal{L}}

\usepackage{times}
\binoppenalty=\maxdimen
\relpenalty=\maxdimen

\title{Efficient Bayesian Inference for a Gaussian Process Density Model}

\author{
{\bf Christian Donner}\thanks{Also affiliated with Bernstein Center for Computational Neuroscience.}  \\
Artificial Intelligence Group \\
Technische Universit\"at Berlin\\
\texttt{christian.donner@bccn-berlin.de} \\
\And
{\bf Manfred Opper}   \\
Artificial Intelligence Group \\
Technische Universit\"at Berlin\\
\texttt{manfred.opper@tu-berlin.de}} 

\begin{document}

\maketitle

\begin{abstract}
We reconsider a nonparametric density model based on Gaussian processes. By augmenting 
the model with latent P\'olya--Gamma random variables and a latent marked Poisson process 
we obtain a new likelihood which is conjugate to the model's Gaussian process prior. The 
augmented posterior allows for efficient inference by Gibbs sampling and an approximate variational mean field approach. For the latter we utilise sparse GP approximations to tackle the infinite dimensionality
of the problem. The performance of both algorithms and comparisons with other density 
estimators are demonstrated on artificial and real datasets with up to several thousand data points.
\end{abstract}

\section{INTRODUCTION}
Gaussian processes (GP) provide highly flexible nonparametric prior distributions over functions~\cite{rasmussen2006gaussian}. They have been 
successfully applied to various statistical problems such as e.g. regression~\cite{williams1996gaussian}, classification~\cite{nickisch2008approximations}, point processes~\cite{adams2009tractable} or the modelling of dynamical systems~\cite{archambeau2007gaussian,damianou2011variational}. Hence, it would seem natural to apply Gaussian processes also to density estimation which is one of the most basic statistical problems. GP density estimation, however, is a nontrivial task: Typical realisations of a GP do not respect non--negativity and normalisation of a probability density. Hence, functions drawn from a GP prior have to be passed through a nonlinear squashing function and the results have to be normalised subsequently to model a density. These operations make the corresponding posterior distributions non--Gaussian. Moreover, likelihoods depend on all the infinitely many GP function values in the domain rather than on the finite number of function values at observed data points. Since analytical inference is impossible, \cite{murray2009gaussian} introduced an interesting Markov chain Monte--Carlo sampler which allows for (asymptotically) exact inference for a Gaussian process density model, where the GP is passed through a sigmoid link function.\footnote{See \cite{riihimaki2014laplace} for an alternative model allowing, however, only for approximate inference schemes.} The approach is able to deal with the infinite dimensionality of the model, because the sampling of the GP variables is reduced to a finite dimensional problem by a point process representation. However, since the likelihood of the GP variables is not conjugate to the prior, the method has to resort to a time--consuming Metropolis--Hastings approach. In this paper we will use recent results on representing the sigmoidal squashing function as an infinite mixture of Gaussians involving P\'olya--Gamma random variables~\cite{polson2013bayesian} to augment the model in such a way that the model becomes tractable by a simpler Gibbs sampler. The new model structure allows also for a much faster variational Bayesian approximation. 

The paper is organised as follows: Sec.~\ref{sec:model} introduces the GP density model, followed by an augmentation scheme that makes its likelihood conjugate to the GP prior. With this model representation we derive two efficient Bayesian inference algorithms in Sec.~\ref{sec:inference}, namely an exact Gibbs sampler and an approximate, fast variational Bayes algorithm. The performance of both algorithms is demonstrated in Sec.~\ref{sec:results} on artificial and real data. Finally, Sec.~\ref{sec:discussion} discusses potential extensions of the model.

\section{GAUSSIAN PROCESS DENSITY MODEL}\label{sec:model}
The generative model proposed by \cite{murray2009gaussian} constructs densities over some $d$-dimensional data space $\X$ 
to be of the form
\begin{equation}\label{eq:density}
\rho(\bx\vert g) = \frac{\sigma(g(\bx))\pi(\bx)}{\int_\X\sigma(g(\bx))\pi(\bx)d\bx}.
\end{equation}
$\pi(\bx)$ defines a (bounded) base probability measure over $\X$, 
which is usually taken from a fixed parametric family. The denominator ensures normalisation $\int_\X \rho(\bx \vert g) d\bx = 1$. The choice of $\pi(\bx)$ is important as will be discussed Sec.~\ref{sec:discussion}. A prior distribution over densities is introduced by assuming 
a Gaussian process prior \cite{rasmussen2006gaussian} over the function $g(\bx):\X\rightarrow\R$. 
The GP is defined by a  mean function $\mu(\bx)$ (in this paper, we consider only constant mean functions $\mu(\bx)=\mu_0$)  and covariance kernel 
$k(\bx,\bx^\prime)$. Finally, $\sigma(z) = \frac{1}{1 + e^{- z}}$ is the sigmoid function, which guarantees that
the density is non--negative and bounded.

In Bayesian inference, the 
posterior distribution  of $g$ given observed data $\dataset = \set{\bx_n}_{n=1}^N$ with $\bx\in \X$ is
computed from the GP prior $p(g)$ and the likelihood as
\begin{equation}
p(g\vert \dataset) \propto p(\dataset\vert g)p(g).
\end{equation}
The likelihood is given by
\begin{equation}\label{eq:likelihood}
p(\dataset\vert g) = \frac{\prod_{n=1}^N\sigma(g(\bx_n))\pi(\bx_n)}{\left(\int_\mathcal{\X}\sigma(g(\bx))\pi(\bx)d\bx\right)^{N}}.
\end{equation}
Practical inference for this problem, however, is non-trivial, because (i) the posterior is non--Gaussian
and (ii) the likelihood involves an integral of $g$ over the whole space. Thus, in contrast to
simpler problems such as GP regression or classification, it is impossible to reduce inference to finite dimensional
integrals. To circumvent the problem that the likelihood is not conjugate to the GP prior, 
 \cite{murray2009gaussian} proposed a Metropolis-Hastings MCMC algorithm for this model.
 We will show in the next sections that one can augment the model with auxiliary latent random variables
in such a way that the resulting likelihood is of a conjugate form  allowing for a more efficient  
Gibbs sampler with explicit conditional probabilities.

\subsection{LIKELIHOOD AUGMENTATION}
To obtain a likelihood which is conjugate to the GP $p(g)$ we require that it assumes a Gaussian form in $g$. \paragraph{Representing the denominator}
As a starting point, we follow \cite{walker2011posterior} and use the
representation
\begin{equation}
\frac{1}{z^{N}} = \frac{\int_0^\infty \lambda^{N-1} e^{-\lambda z}d\lambda}{\Gamma(N)},
\end{equation}
where $\Gamma(\cdot)$ is the gamma function. Identifying $z=\int_\mathcal{\X}\sigma(g(\bx))\pi(\bx)d\bx$ in Eq.~\eqref{eq:likelihood} we can rewrite the likelihood as
$p(\dataset\vert g) = \int_0^\infty p(\dataset,\lambda\vert g) d\lambda$ where
\begin{equation}\label{eq:augmenation1}
\begin{split}
p(\dataset,\lambda\vert g) \propto &  \exp\left(-\int_\X\lambda\sigma(g(\bx))\pi(\bx)d\bx\right) \\
& \times p(\lambda)\; \prod_{n=1}^N\lambda \sigma(g(\bx_n))\pi(\bx_n),
\end{split}
\end{equation}
with the improper prior $p(\lambda)=\lambda^{-1}$ over the auxiliary latent variable $\lambda$.
To transform the likelihood further into a form which is Gaussian in $g$, we utilise a
representation of the sigmoid function as a scale mixture of Gaussians.
\paragraph{P\'olya--Gamma representation of sigmoid function}
As discovered by \cite{polson2013bayesian}, the inverse hyperbolic cosine 
can be represented as an infinite mixture of scaled Gaussians
\begin{equation}\label{eq:polya-gamma}
\cosh^{-b}(z/2)=\int_0^\infty e^{-\frac{z^2}{2}\omega}\PG(\omega\vert b,0) d\omega,
\end{equation}
where $\PG(\omega\vert b,0)$ is the {\it P\'olya--Gamma density} of random variable $\omega\in \R^+$. Moments of those densities can be easily computed~\cite{polson2013bayesian}.
Later, we will also use the {\it tilted} P\'olya-Gamma densities defined as
\begin{equation}\label{eq:tilted PG}
\PG(\omega\vert b, c)\propto \exp\left(-\frac{c^2}{2}\omega\right)\PG(\omega\vert b,0).
\end{equation}

These definitions allows for a Gaussian representation of the sigmoid function as
\begin{equation}\label{eq:PG sigmoid}
\sigma(z)=\frac{e^{z/2}}{2\cosh(z/2)} = \int_0^\infty e^{f(\omega,z)}\PG(\omega\vert 1,0)d\omega
\end{equation}
with $f(\omega,z)=\frac{z}{2}-\frac{z^2}{2}\omega - \ln 2$.  This result will be used to transform 
the products over observations $\sigma(g(\bx_n))$ in the likelihood \eqref{eq:augmenation1}
into a Gaussian form.

We will next deal with the first term in the likelihood \eqref{eq:augmenation1}
which contains the integral  over $\bx$. For this part of the model we will derive a point process
representation which can be understood as a generalisation of the approach of \cite{murray2009gaussian}.
\paragraph{Marked--Poisson representation}
Utilising the sigmoid property $\sigma(z) = 1 - \sigma(-z)$ 
and the P\'olya-Gamma representation \eqref{eq:PG sigmoid}
the integral in the exponent of Eq.~\eqref{eq:augmenation1} can be written as a double integral
\begin{equation}
\label{eq:exponent}
\begin{split}
& -\int_\X\lambda\sigma(g(\bx))\pi(\bx)d\bx = \\
& \int_\X(\sigma(-g(\bx)) - 1)\lambda\pi(\bx)d\bx = \\
& \int_\X \int_{\R^+}  \left(e^{f(\omega, - g(\bx))} - 1\right) \lambda\pi(\bx)\PG(\omega\vert 1,0)
d\omega d\bx
\end{split}
\end{equation}
Next we will use a result for the characteristic function of a Poisson process. Following~\cite[chap. 3]{kingman1993poisson} one has 
\begin{equation}\label{eq:characteristic}
\EE{\phi}{\prod_{\bz\in\Pi}h(\bz)} = \exp \left(\int_{\Z}(h(\bz) - 1)\phi(\bz) d\bz\right).
\end{equation}
$h(\cdot)$ is a function on a space $\Z$ and the expectation is over a Poisson process
$\Pi$ with rate function $\phi(\bz)$. $\Pi=\set{\bz_m}_{m=1}^M$ denotes a random set of points on the space $\Z$. 
To apply this result to our problem, we identify $\Z = \X \times R^+$, $\bz = (\bx,\omega)$ and 
$\phi_\lambda(\bx,\omega)=\lambda\pi(\bx)\PG(\omega\vert 1, 0)$
and finally 
$h(\bz)= e^{f(\omega, - g(\bx))}$ to rewrite the exponential in Eq.~\eqref{eq:augmenation1} as 
\begin{equation}\label{eq:exponent Poisson}
\begin{split}
e^{-\int_\X\lambda\sigma(g(\bx))\pi(\bx)d\bx}  = \small \EE{{\phi}_{\lambda}}{\prod_{(\omega,\bx)\in \Pi} e^{f(\omega,-g(\bx))}}.
\end{split}
\end{equation}
By substituting Eq.~\eqref{eq:PG sigmoid} and~\eqref{eq:exponent Poisson} into Eq.\eqref{eq:augmenation1} we obtain the final augmented form of the likelihood of Eq.~\eqref{eq:likelihood} which is one of the main results of our paper.
\begin{equation}\label{eq:augmented likelihood}
\begin{split}
p(\dataset, \lambda,\Pi ,&\bomega_N \vert g) \propto \prod_{n=1}^N \phi_\lambda(\bx_n, \omega_n) e^{f(\omega_n,g(\bx_n))} \\
& \times p_{\phi_\lambda}(\Pi\vert\lambda)p(\lambda)\; \prod_{(\omega,\bx)\in\Pi} e^{f(\omega,-g(\bx))},
\end{split}
\end{equation}
with $p_{\phi}(\Pi\vert\lambda)$ being the density over a Poisson process $\Pi=\set{(\bx_m,\omega_m)}_{m=1}^M$ in the augmented space $\X \times \R^+$  with intensity   
$\phi_\lambda(\bx,\omega)$. 
\footnote{Densities such as $p_{\phi_\lambda}(\Pi\vert\lambda)$ could be understood
as the Radon--Nykodym derivative~\cite{konstantopoulos2011radon} of the corresponding probability measure 
with respect to some fixed dominating measure. However, we will not need an explicit form here.}
This new process can be identified as a {\it marked Poisson process} \cite[chap. 5]{kingman1993poisson}, where the events $\set{\bx_m}_{m=1}^M$ in the original data space $\X$ follow a Poisson process with rate $\lambda\pi(\bx)$. Then, on each event $\bx_m$ an independent {\it mark} $\omega_m\sim\PG(\omega_m\vert b,0)$ is drawn at random from the P\'olya--Gamma density.
Finally, $\bomega_N=\set{\omega_n}_{n=1}^N$ is the set of latent P\'olya--Gamma variables which result from the sigmoid augmentation at the observations $\bx_n$.

\paragraph{Augmented posterior over GP density}
With Eq.~\eqref{eq:augmented likelihood} we obtain the joint posterior over the GP $g$, the rate scaling $\lambda$, the marked Poisson process $\Pi$, and the P\'olya--Gamma variables at the observations $\bomega_N$ as
\begin{equation}\label{eq:joint posterior}
p(\bomega_N, \Pi , \lambda, g\vert \dataset)\propto p(\dataset, \bomega_N, \Pi , \lambda \vert g)p(g).
\end{equation}
In the following, this new representation will be used to derive two inference algorithms.
\section{INFERENCE}\label{sec:inference}
We will first derive an efficient  Gibbs sampler which (asymptotically) solves the inference problem exactly, and then a variational mean-field algorithm, which only finds an approximate solution, but in a much faster time.
\subsection{GIBBS SAMPLER}
Gibbs sampling~\cite{geman1987stochastic} generates samples from the posterior by creating a Markov chain, where at each time, a block of variables is drawn from the conditional posterior given all the other variables. Hence, to perform Gibbs sampling, we have to derive these conditional distributions for each set of variables from Eq.~\eqref{eq:joint posterior}. Most of the following results are easily obtained by direct inspection. The only non--trivial case is the conditional distribution over the latent point process $\Pi$.

\paragraph{P\'olya-Gamma variables at observations} 
The conditional posterior over the set of P\'olya--Gamma variables $\bomega_N$ depends only on the function $g$ at the observations $\set{g(\bx_n)}_{n=1}^N$ and turns out to be 
\begin{equation}\label{eq:PG observation}
p(\bomega_N\vert \bg) = \prod_{n=1}^N\PG(\omega_n\vert 1, g(\bx_n)),
\end{equation}
where we have used the definition of a tilted P\'olya-Gamma density in Eq.~\eqref{eq:tilted PG}.
This density can be efficiently sampled by methods developed by~\cite{polson2013bayesian}\footnote{The sampler implemented by~\cite{linderman2017git} is used for this work.}.

\paragraph{Rate scaling} The rate scaling $\lambda$ has a conditional  Gamma density
given by
\begin{equation}\label{eq:rate scaling gibbs}
\rm{Gamma}(\lambda\vert \alpha, 1) = \frac{(\lambda)^{\alpha-1}e^{-\lambda}}{\Gamma(\alpha)}.
\end{equation}
with $\alpha=|\Pi| + N= M + N$. Hence, the posterior is dependent on the number of observations and the number on events of the marked Poisson process $\Pi$.

\paragraph{Posterior Gaussian process} Due to the form of the augmented likelihood the conditional posterior for the GP $\bg_{N+M}$ at the observations $\set{\bx_n}_{n=1}^N$ and the latent events $\set{\bx_m}_{m=1}^M$ is a
multivariate Gaussian density
\begin{equation}\label{eq:GP gibbs}
p(\bg_{N+M}\vert \Pi,\bomega_N) = \mathcal{N}(\bmu_{N+M},\Sigma_{N+M}),
\end{equation}
with covariance matrix  $\Sigma_{N+M} = [D + K_{N+M}^{-1}]^{-1}$. The diagonal matrix $D$ has its first $N$ entries given by $\bomega_N$ followed by $M$ entries being $\set{\omega_m}_{m=1}^M$. The mean is $\bmu_{N+M} = \Sigma_{N+M} \left[\bs{u} + K_{N+M}^{-1}\bmu_0^{(N+M)}\right]$, where the first $N$ entries of $N+M$ dimensional vector $\bs{u}$ are $1/2$ and the rest are $-1/2$. $K_{N+M}$ is the prior covariance kernel matrix of the GP evaluated at 
the observed points $\bx_n$ and the latent events $\bx_m$, and $\bmu_0^{(N+M)}$ is an $N+M$ dimensional vector with all entries being $\mu_0$. 

The predictive conditional posterior for the GP for any set of points in $\X$ is simply given via the conditional prior $p(g\vert \bg_{N+M})$, which has a well known form and can be found in~\cite{rasmussen2006gaussian}.

\paragraph{Sampling the latent marked point process} 
We easily find that the conditional posterior of the marked point process is given by
\begin{align}\label{eq:latent poisson gibbs}
\scriptstyle p(\Pi\vert g, \lambda) = & \scriptstyle \frac{\prod_{\omega,\bx\in \Pi} e^{f(\omega,-g(\bx))}
p_{\phi_\lambda}(\Pi\vert\lambda)}{\exp\left(\int_{\X\times \R^+} \left(e^{f(\omega,-g(\bx))}-1\right)
\phi_\lambda(\bx,\omega) d\omega d\bx \right)}, &
\end{align}
where the form of the normalising denominator is obtained using Eq.~\eqref{eq:characteristic}. 
By computing the characteristic function of this conditional point process (see App.~\ref{app:poisson proof}) we can show that it is again a marked Poisson process with intensity
\begin{equation}\label{eq:gibbs posterior rate}
\Lambda(\bx,\omega) = \lambda \pi(\bx) \sigma(-g(\bx))\PG(\omega\vert 1, g(\bx)).
\end{equation} 
To sample from this process we first draw Poisson events $\bx_m$ in the original data space $\X$ using the rate $\int_{\R^+} \Lambda(\bx,\omega) d\omega = \lambda \pi(\bx) \sigma(-g(\bx))$ \cite[chap. 5]{kingman1993poisson}. Subsequently for each event $\bx_m$ 
a mark $\omega_m$ is generated from the conditional density $\omega_m\sim \PG(\omega\vert 1, g(\bx_m))$.

To sample the events $\set{\bx_m}_{m=1}^M$, we use the well known approach of {\em thinning}
\cite{adams2009tractable}. We note, that the rate is upper bounded by the base measure $\lambda \pi(\bx)$. 
Hence, we first generate points $\tilde{\bx}_m$ from a Poisson process with intensity
$\lambda \pi(\bx)$. This is easily achieved by noting that 
the required number $M_{\rm max}$ of such events is Poisson distributed with mean parameter $\int_\X\lambda \pi(\bx)dx = \lambda$. The position of the events can then be obtained by 
sampling $\set{\tilde{\bx}_m}_{m=1}^{M_{\rm max}}$ independent points from  the base density $\tilde{\bx}_m\sim\pi(\bx)$. These events are {\it thinned} by keeping each point $\tilde{\bx}_m$ with probability  
$\sigma(-g(\tilde{\bx}_m))$. The kept events constitute the final set $\set{\bx_m}_{m=1}^M$.

\paragraph{Sampling hyperparameters} In this work we will consider specific functional forms for the kernel
$k(\bx,\bx^\prime)$ and the base measure $\pi(\bx)$ which are parametrised by hyperparameters $\bs{\theta}_k$ and $\bs{\theta}_\pi$. These will be sampled by a Metropolis-Hastings method~\cite{hastings1970monte}. The GP prior mean 
$\mu_0$ can be directly sampled from the conditional posterior given $\bg_{M+N}$. In this work, the hyperparameters are sampled every $v=10$ step. Different choices of $v$ might yield faster convergence of the Markov Chain.
Pseudo code for the Gibbs sampler is provided in Alg.~\ref{alg:Gibbs}.

\begin{algorithm}\label{alg:Gibbs}
    \SetKwInOut{Init}{Init}

    \Init{$\set{\bx_m}_{m=1}^M$, $\bg_{N+M}$, $\lambda$, and $\btheta_k$, $\btheta_\pi$, $\mu_0$}
    \For{\text{Length of Markov chain}}
      {
      	{\bf Sample PG variables at $\set{\bx_m}$}: $\bomega_N\sim$ Eq.~\eqref{eq:PG observation}\\
        {\bf Sample latent Poisson process}: $\Pi\sim$ Eq.~\eqref{eq:latent poisson gibbs}\\
        {\bf Sample rate scaling}: $\lambda\sim$ Eq.~\eqref{eq:rate scaling gibbs}\\
        {\bf Sample GP}: $\bg_{N+M}\sim$ Eq.~\eqref{eq:GP gibbs}\\
        {\bf Sample hyperparameters}: Every $v^{\rm th}$ sample with Metropolis--Hastings\\
      }
      
    \caption{Gibbs sampler for GP density model.}
\end{algorithm}
\subsection{VARIATIONAL BAYES}
While expected to be more efficient than a Metropolis-Hastings sampler 
based on the unaugmented likelihood~\cite{murray2009gaussian}, the Gibbs sampler is practically still limited. The main computational bottleneck comes from the sampling of the conditional Gaussian over function values of $g$.  The computation of the covariances requires the 
inversion of matrices of dimensions $N+M$, with a complexity $\mathcal{O}((N+M)^3)$. Hence the algorithm does not only become infeasible, when we have many observations, i.e when $N$ is large, but also if the sampler requires many thinned events, i.e. if $M$ is large. This can happen in particular for bad choices of the base measure $\pi(\bx)$. 
In the following, we introduce a variational Bayes algorithm~\cite{bishop2006pattern}, which solves the inference 
problem approximately, but with a complexity which scales linearly in the data size and is independent of structure.

\paragraph{Structured mean--field approach} The idea of variational inference~\cite{bishop2006pattern} is
to approximate an intractable posterior $p(Z\vert\dataset)$ by a simpler distribution $q(Z)$ from a tractable family.  $q(Z)$ is optimised  by minimising the Kullback-Leibler divergence between $q(Z)$ and $p(Z\vert\dataset)$
which is equivalent  to maximising the so called {\it variational lower bound} (sometimes also called ELBO for evidence lower bound) given by
\begin{equation}\label{eq:lower bound}
\LB(q(Z)) = \EE{Q}{\ln \frac{p(Z,\dataset)}{q(Z)}} \leq \ln p(\dataset),
\end{equation}
where $Q$ denotes the probability measure with density $q(Z)$. 
A common approach for variational inference is a structured mean--field method, where dependencies between sets of variables are neglected. For the problem at hand we assume that
\begin{equation}\label{eq:variational posterior}
q(\bomega_N,\Pi,g,\lambda) = q_1(\bomega_N,\Pi)q_2(g,\lambda).
\end{equation}
A standard result for the variational mean--field approach shows that the optimal independent factors, which maximise the lower bound in Eq.~\eqref{eq:lower bound} are given by
\begin{equation}\label{eq:optimal factor1}
\ln q_1(\bomega_N,\Pi) = \EE{Q_2}{\ln p(\dataset, \bomega_N, \Pi, \lambda, g)} + {\rm const.},
\end{equation}
\begin{equation}\label{eq:optimal factor2}
\ln q_2(g,\lambda) = \EE{Q_1}{\ln p(\dataset, \bomega_N, \Pi, \lambda, g)} + {\rm const}.
\end{equation}
By inspecting Eq.~\eqref{eq:joint posterior},~\eqref{eq:optimal factor1}, and ~\eqref{eq:optimal factor2} it turns out that the densities of all four sets of variables factorise as
\begin{align}
& q_1(\bomega_N,\Pi) = q_1(\bomega_N)q_1(\Pi), \\
& q_2(g,\lambda) = q_2(g)q_2(\lambda).
\end{align}
We will optimise the  factors by a straightforward iterative algorithm, where each factor is updated given expectations over the others based on the previous step. Hence, the lower bound in Eq.~\eqref{eq:lower bound} is increased in 
each step. Again we will see that the augmented likelihood in Eq.~\eqref{eq:augmented likelihood} allows for analytic solutions of all required factors.

\paragraph{P\'olya--Gamma variables at the observations} Similar to the Gibbs sampler, the variational posterior of the P\'olya-Gamma variables at the observations is a product of tilted P\'olya--Gamma densities given by
\begin{equation}\label{eq: PG observations VB}
q_1(\bomega_N) = \prod_{n=1}^N \PG(\omega_n\vert 1, c_n),
\end{equation}
with $c_n=\sqrt{\EE{Q_2}{g(\bx_n)^2}}$. The only difference is, that the second argument of $\PG$ depends on the expectation of the square of $g(\bx_n)$.

\paragraph{Posterior marked Poisson process} Similar to the corresponding result for the Gibbs sampler
we can show\footnote{The proof is similar to the one from App.~\ref{app:poisson proof}.} that the optimal latent point process $\Pi$ is a Poisson process
with rate given by
\begin{equation}\label{eq:poisson rate VB}
\begin{split}
\Lambda_1(\bx,\omega)=& \lambda_1\pi(\bx)\sigma(-c(\bx))\PG(\omega\vert 1, c(\bx)) \\
& \times e^{\left(c(x) - g_1(\bx)\right)/2}
\end{split}
\end{equation}
with  $\lambda_1=e^{\EE{Q_2}{\ln \lambda}}$, $c(\bx)=\sqrt{\EE{Q_2}{f(\bx)^2}}$, and $g_1(\bx)=\EE{Q_2}{g(x)}$. Note also the similarity to the Gibbs sampler in Eq.~\eqref{eq:gibbs posterior rate}.

\paragraph{Optimal posterior for rate scaling} The posterior for the rate scaling $\lambda$ is a Gamma distribution
given by
\begin{equation}\label{eq:optimal lambda density}
q_2(\lambda) = \rm{Gamma}(\lambda\vert \alpha_{2}, 1) = \frac{\lambda^{\alpha_{2}-1}e^{-\lambda}}{\Gamma(\alpha_{2})},
\end{equation}
where $\alpha_{2}= N + \EE{Q_1}{\sum_{\bx^\prime\in\Pi}\delta(\bx - \bx^\prime)}$, and $\EE{Q_1}{\sum_{\bx^\prime\in\Pi}\delta(\bx - \bx^\prime)} = \int_\X\int_{\R^+} \Lambda_1(\bx,\omega) d\omega d\bx$, and $\delta(\cdot)$ is the Dirac delta function. The integral is solved by importance sampling as will be explained (see Eq.~\eqref{eq:importance sampling}).

\paragraph{Approximation of GP via sparse GP} The optimal variational form for the posterior  $g$ is a GP
given by \begin{equation}\label{eq:unsparse posterior}
q_2(g) \propto e^{U(g)}p(g),
\end{equation}
where $U(g)=\EE{Q_1}{\ln p(\dataset,\bomega_N, \Pi , \lambda \vert g)}$ results in the Gaussian 
log--likelihood
\begin{equation}
U(g) = -\frac{1}{2}\int_\X A(\bx)g(\bx)^2 d\bx + \int_\X B(\bx)g(\bx)d\bx + {\rm const.}
\end{equation}
with
\begin{equation}
A(\bx) = \sum_{n=1}^N \EE{Q_1}{\omega_n}\delta(\bx - \bx_n) + \int_{\R^+} \omega \Lambda_{1}(\bx,\omega) d\omega ,
\end{equation}
\begin{equation}
B(\bx) = \frac{1}{2} \sum_{n=1}^N \delta(\bx - \bx_n) - \frac{1}{2}\int_{\R^+} \Lambda_1(\bx,\omega) 
d\omega.
\end{equation}
For general GP priors, this free form optimum is intractable by the fact that the likelihood
depends on $g$ at infinitely many points. Hence, we resort to an additional approximation which makes the dimensionality of the problem again finite. The well known framework of {\it sparse} GPs \cite{csato2002phd,csato2002tap,titsias2009variational} turns out to be useful in this case. This has been 
introduced for likelihoods with large, but finite dimensional likelihoods~\cite{titsias2009variational,snelson2006sparse} and later generalised to infinite dimensional problems \cite{matthews2016sparse,batz2017approximate}. 
The sparse approximation assumes a variational posterior of the form
\begin{equation}\label{eq:full sparse posterior}
q_2(g) = p(g\vert\gs)q_2(\gs),
\end{equation}
where $\gs$ is the GP evaluated at a finite set of {\it inducing points} $\set{\bx_l}_{l=1}^L$ and $p(g\vert\gs)$ is the conditional prior. A variational optimisation yields
\begin{equation}\label{eq:sparse posterior}
q_2(\gs) \propto e^{U^s(\gs)}p(\gs),
\end{equation}
where the first term can be seen as a new `effective' likelihood only depending on the inducing points. This new (log) likelihood is given by
\begin{equation}
\begin{split}
& U^s(\gs) = \EE{P}{U(g)\vert \gs} = \\
& -\frac{1}{2}\int_\X A(\bx)\tilde{g}_s(\bx)^2 d\bx + \int_\X B(\bx)\tilde{g}_s(\bx)d\bx + {\rm const.},
\end{split}
\end{equation}
with $\tilde{g}_s(\bx) = \mu_0 + \boldsymbol{k}_s(\bx)^\top\; K_s^{-1}(\boldsymbol{g}_s - \bmu_0^{(L)})$, $\boldsymbol{k}_s(\bx)$ being an $L$ dimensional vector, where the $l^{\rm th}$ entry is $k(\bx,\bx_l)$ and $K_s$ being the prior covariance matrix for all inducing points. The expectation is computed with respect to the GP prior conditioned on the sparse GP $\gs$. We identify Eq.~\eqref{eq:sparse posterior} being a multivariate normal distribution with covariance matrix
\begin{equation}\label{eq:posterior GP cov}
\Sigma_2^s = \left[K_s^{-1}\int_{\X} A(\bx)\bs{k}_s(\bx)^\top\bs{k}_s(\bx) d\bx\; K_s^{-1} + K_s^{-1}\right]^{-1},
\end{equation}
and mean
\begin{equation}\label{eq:posterior GP mean}
\boldsymbol{\mu}_2^s = \Sigma_2^s\left(K_s^{-1}\int_{\X} \bs{k}_s(\bx) \tilde{B}(\bx)d\bx + K_s^{-1}\bmu_0^{(L)}\right),
\end{equation}
with $\tilde{B}(\bx) = B(\bx) - A(x)(\mu_0 - \bs{k}_s(\bx)^\top K_s^{-1}\bmu_0^{(L)})$. 
\paragraph{Integrals over $\bx$} 
The sparse GP approximation and the posterior over $\lambda$ in Eq.~\eqref{eq:optimal lambda density} requires the computation of integrals of the form
\begin{equation}
I \doteq \int_\X \int_{\R^+} y(\bx,\omega) \Lambda_1(\bx,\omega)d\omega d\bx,
\end{equation}
with specific functions $y(\bx,\omega)$. For these functions, the inner integral over $\omega$
can be computed analytically, but the outer one over the space $\X$ has to be treated
numerically. We approximate it via importance sampling 
\begin{equation}\label{eq:importance sampling}
I  \approx \frac{1}{R}\sum_{r=1}^R \int_{\R^+} y(\bx_r,\omega_r) \frac{\Lambda_1(\bx_r,\omega_r)}{\pi(\bx_r)}d\omega_r,
\end{equation}
where every sample point $\bx_r$ is independently drawn from the base measure $\pi(\bx)$.

\paragraph{Updating hyperparameters} Having an analytic solution for every factor of the variational posterior in Eq.~\eqref{eq:variational posterior} we further require the optimisation of hyperparameters. $\btheta_k$, $\btheta_\pi$ and $\mu_0$ are optimised by maximising the lower bound in Eq.~\eqref{eq:lower bound} (see App.~\ref{app:LB} for explicit form) with a gradient ascent algorithm having an adaptive learning rate (Adam) \cite{kingma2014adam}. Additional hyperparameters are the locations of inducing points $\set{\bx_l}_{l=1}^L$. Half of them are drawn randomly from the initial base measure, while half of them are positioned on regions with a high density of observations found by a k--means algorithm. Pseudo code for the complete variational algorithm is provided in Alg.~\ref{alg:VB}.

\begin{algorithm}[t]\label{alg:VB}
    \SetKwInOut{Init}{Init}
    \SetKw{Updatehyper}{Update $\btheta_k,\btheta_\pi, \mu_0$ with gradient update}
    \SetKwBlock{Updateqone}{Update $q_1$}{}
    \SetKwBlock{Updateqtwo}{Update $q_2$}{}

    \Init{Inducing points, $q_2(\gs)$, $q_2(\lambda)$, and $\btheta_k,\btheta_\pi, \mu_0$}
    \While{$\mathcal{L}$ not converged}
      {
        \Updateqone{
        {\bf PG distributions at observations}: $q_1^*(\bomega_N)$ with Eq.~\eqref{eq: PG observations VB}\\
        {\bf Rate of latent process}: $\Lambda_1(\bx,\omega)$ with Eq.~\eqref{eq:poisson rate VB}       
      		       }
        \Updateqtwo{
        {\bf Rate scaling}: $\alpha_2$ with Eq.~\eqref{eq:optimal lambda density} \\
      {\bf Sparse GP}: $\Sigma_2^s,\mu_2^s$ with Eq.~\eqref{eq:posterior GP cov},~\eqref{eq:posterior GP mean}
      }
      \Updatehyper
      }
      
    \caption{Variational Bayes algorithm for GP density model}
\end{algorithm}

Python code for Alg.~\ref{alg:Gibbs} and~\ref{alg:VB} is provided at \cite{donner2018git}.

\section{RESULTS}\label{sec:results}
To test our two inference algorithms, the Gibbs sampler and the variational Bayes algorithm (VB), we will first evaluate them on data drawn from the generative model. Then we compare both on an artificial dataset and several real datasets. We will only consider cases with $\X=\R^d$. To evaluate the quality of inference we consider always the logarithm of the expected test likelihood
\begin{equation}\label{eq:test likelihood}
\ell_{\rm test}(\tilde{\dataset}) \doteq \ln\left( \EE{}{\prod_{\bx\in \tilde{\dataset}}\rho(\bx)}\right),
\end{equation}
where $\tilde{\dataset}$ is test data unknown to the inference algorithm and the expectation is over the inferred posterior measure. In practice we sample this expectation from the inferred posterior over $g$. Since this quantity involves an integral, that is again approximated by Eq.~\eqref{eq:importance sampling}, we check that the standard deviation ${\rm std}(I)$ is less than $1\%$ of the value of the estimated value $I$.
\paragraph{Data from generative model.} We generate datasets according to Eq.~\eqref{eq:density}, where $g$ is drawn from the GP prior with $\mu_0=0$. As covariance kernel we assume a squared exponential throughout this work
\begin{equation}
k(\bx,\bx^\prime) = \theta_k^{(0)}\prod_{i=1}^d \exp\left(-\frac{(x_i-x_i^\prime)^2}{2(\theta_k^{(i)})^2}\right).
\end{equation}
The base measure $\pi(\bx)$ is a standard normal density. We use the algorithm described in~\cite{murray2009gaussian} to generate exact samples. In this section, the hyperparameters $\btheta_k,\btheta_\pi$ and $\mu_0$ are fixed to the true values for inference. Unless stated otherwise for the VB the number of inducing points is fixed to $200$ and the number of integration points for importance sampling to $5\times 10^3$. For the Gibbs sampler, we sample a Markov chain of $5\times 10^3$ samples after a burn--in period of $2\times 10^3$ samples.
\begin{figure}[t]
\centering
\includegraphics[width=.45\textwidth]{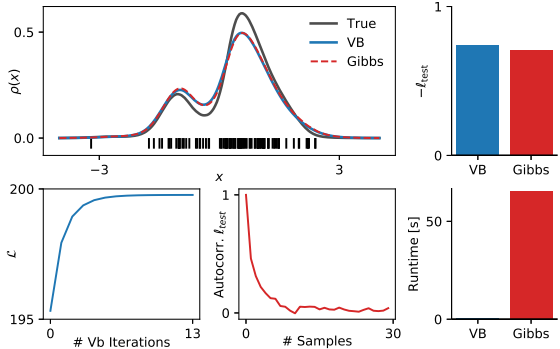}
\caption{{\bf 1D data from the generative model.} Data consist of $100$ samples from the underlying density sampled from the GP density model. {\bf Upper left:} True density (black line), data (black vertical bars), mean posterior density inferred by Gibbs sampler (red dashed line) and VB algorithm (blue line). {\bf Upper right:} Negative log expected test likelihood of Gibbs and VB inferred posterior. {\bf Lower left:} Variational lower bound as function of iterations of the VB algorithm. {\bf Lower middle:} Autocorrelation of test likelihood as function of Markov chain samples obtained from Gibbs sampler. {\bf Lower right:} Runtime of the two algorithms (VB took $0.3\ {\rm s}$).}\label{fig:fig1}
\end{figure}

In Fig.~\ref{fig:fig1} we see a 1 dimensional example dataset, where both inference algorithms recover well the structure of the underlying density. The inferred posterior means are barely distinguishable. However, evaluating the inferred densities on an unseen test set, we note that the Gibbs sampler performs slightly better. Of course, this is expected since the sampler provides exact inference for the generative model and should (on average) not be outperformed by the approximate VB as long as the sampled Markov chain is long enough. In Fig.~\ref{fig:fig1} (bottom left) we see that only $13$ iterations of the VB are required to meet the convergence criterion. For Markov chain samplers to be efficient, correlations between samples should decay quickly. Fig.~\ref{fig:fig1} (bottom middle) shows the autocorrelation of $\ell_{\rm test}$, which was evaluated at each sample of the Markov chain. After about $10$ samples the correlations reach a plateau close to $0$, demonstrating  excellent mixing properties of the sampler. Comparing the run time of both algorithms, VB ($0.3\ {\rm s}$) outperforms the sampler $\sim 1\ {\rm min}$ by more than $2$ orders of magnitude.
 
\begin{figure}[t]
\centering
\includegraphics[width=.45\textwidth]{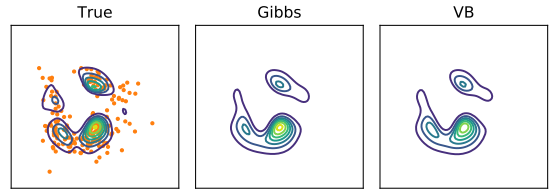}
\caption{{\bf 2D data from generative model.} {\bf Right:} $200$ samples from the underlying two dimensional density. {\bf Middle:} Posterior mean of Gibbs sampler inferred density. {\bf Right:} Posterior mean of VB inferred density.}\label{fig:fig2}
\end{figure}

To demonstrate the inference for more complicated problems, $2$ dimensional data are generated with $200$ samples (Fig.~\ref{fig:fig2}). The posterior mean densities inferred by both algorithms capture the structure well. As before, the log expected test likelihood is larger for the Gibbs sampler ($\ell_{\rm test} = -296.2$) compared to VB ($\ell_{\rm test} = -306.0$). However, the Gibbs sampler took $>20\ {\rm min}$ while the VB required only $1.8\ \rm{s}$ to obtain the result.

\begin{table}[t]
\centering
\begin{tabular}{cc|cc|cc}
Dim & \# points & \multicolumn{2}{c|}{Gibbs} & \multicolumn{2}{c}{VB} \\
    &                &  $\ell_{\rm test}$         & $T$ {[}s{]}  &   $\ell_{\rm test}$      & $T$ {[}s{]} \\ \hline
1   & 50             & -146.9   & 30.1          & -149.2 & 1.13         \\
2   & 100            & -257.0   & 649.9         & -260.2 & 2.03         \\
2   & 200            & -285.3   & 546.1         & -289.6 & 1.41         \\
6   & 400            & -823.9   & 4667          & -822.2 & 0.89        
\end{tabular}
\caption{{\bf Performance of Gibbs sampler and VB} on different datasets sampled from generative model. $\ell_{\rm test}$ was evaluated on a unknown test set including $50$ samples. In addition, runtime $T$ is reported in seconds.}
\label{tab:tab1}
\end{table}

In Tab.~\ref{tab:tab1} we show results for datasets with different size and different dimensionality. 
The results confirm that the run time  for the Gibbs sampler scales strongly with size and dimensionality of a
problem, while the VB algorithm seems relatively unaffected in this regard. However, the VB is in general outperformed by the sampler in terms of expected test likelihood or in the same range. Note, that the runtime of the Gibbs sampler does not solely depend on the number of observed data points $N$ (compare data set 2 and 3 in Tab.~\ref{tab:tab1}). As discussed earlier this can happen, when the base measure $\pi(\bx)$ is very different from the target density $\rho(\bx)$ resulting in many latent Poisson events (i.e. $M$ is large).

\begin{figure}[t]
\centering
\includegraphics[width=.45\textwidth]{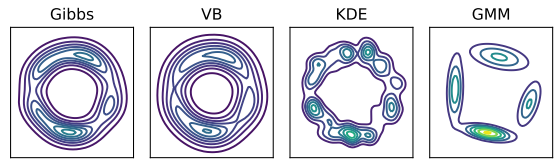}
\caption{{\bf Comparison to other density estimation methods on artificial 2D data.} Training data consist of $100$ data points uniformly distributed on a circle ($1.5$ radius) and additional Gaussian noise ($0.2$ std.). {\bf From left to right:} The posterior mean inferred by Gibbs sampler and VB algorithm, followed by density estimation using KDE and GMM.}\label{fig:fig3}
\end{figure}

\begin{table}[t]
\centering
\begin{tabular}{l|cccc}
                  & Gibbs   & VB      & KDE     & GMM     \\ \hline
$\ell_{\rm test}$ & -220.31 & -230.53 & -228.43 & -237.34
\end{tabular}
\caption{{\bf Log expected test likelihood for circle data.}}
\label{tab:tab2}
\end{table}

\paragraph{Circle data} In the following, we compare the GP density model and its two inference algorithms with two alternative density estimation methods. These are given by a kernel density estimator (KDE) with a Gaussian kernel and a Gaussian mixture model (GMM) \cite{scikit2011learn}. The free parameters of these models (kernel bandwidth for KDE and number of components for GMM) are optimised by $10$-fold cross--validation. Furthermore, GMM is initialised $10$ times and the best result is reported. For the GP density model a Gaussian density is assumed as base measure $\pi(\bx)$, and hyperparameters $\btheta_\pi,\ \btheta_k$, and $\mu_0$ are now optimised. Similar to~\cite{murray2009gaussian} we consider $100$ samples uniformly drawn from a circle with additional Gaussian noise. The inferred densities (only the mean of the posterior for Gibbs and VB) are shown in Fig.~\ref{fig:fig3}. Both GP density methods recover well the structure of the data, but the VB seems to overestimate the width of the Gaussian noise compared to the Gibbs sampler. While the KDE also recovers relatively well the data structure the GMM fails in this case. This is also reflected on the log expected test likelihoods (Tab.~\ref{tab:tab2}).

\begin{figure}[t]
\centering
\includegraphics[width=.45\textwidth]{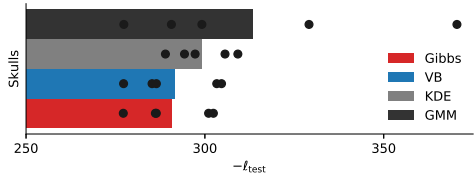}
\caption{{\bf Performance on `Egyptian Skulls' dataset~\cite{hand1993handbook}.} $100$ training points and $4$ dimensions. Bar height shows average negative log test likelihood obtained by five random permutations of training and test set and points mark single permutation results.}\label{fig:fig4}
\end{figure}
\paragraph{Real data sets} The `Egyptian Skulls' dataset~\cite{hand1993handbook} contains $150$ data points in $4$ dimensions. $100$ training points are randomly selected and performance is evaluated on the remaining ones. Before fitting data is whitened. Base measure and fitting procedure for all algorithms are the same as for the circular data. Furthermore, fitting is done for $5$ random permutations of training and test set. The results in Fig.~\ref{fig:fig4} show that both algorithms for the GP density model outperform the two other ones on this dataset.

\begin{figure}[t]
\centering
\includegraphics[width=.45\textwidth]{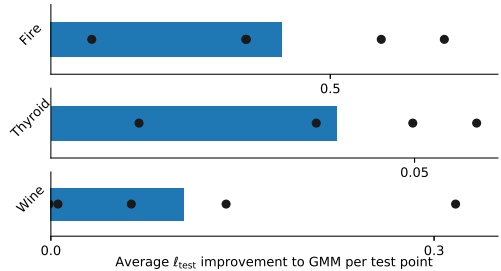}
\caption{{\bf Application on higher dimensional data with many data points.} The improvement on log expected test likelihood $\ell_{\rm test}$ per test point compared to GMM, when using same as base measure $\pi(\bx)$ for the VB inference. {\bf From top to bottom:} `Forest Fire' dataset ~\cite{dua2017uci,cortez2007data} ($400$ training points, $117$ test points, $5$ dim.), `Thyroid' dataset \cite{keller2012hics} ($3\times10^3$, $772$, $6$), `Wine' dataset~\cite{dua2017uci} ($6\times10^3$, $498$, $9$). Bars mark improvement on average of random permutations of training and test set while points mark single runs.}\label{fig:fig5}
\end{figure}

Often practical problems may consist of many more data points and dimensions. As discussed, the Gibbs sampler is not practical for such kind of problems, while the VB could handle larger amounts of data. Unfortunately, the sparsity assumption and the integration via importance sampling is expected to become poorer with increasing number of dimensions. Noting, however, that the `effective' dimensionality in our model is determined by the base measure $\pi(\bx)$, one can circumvent this problem by an educated choice of $\pi(\bx)$ if data $\dataset$ lie in a submanifold of the high dimensional space $\X$.

We employ this strategy by first fitting a GMM to the problem and then utilising the fit as base measure. In Fig.~\ref{fig:fig5} we consider $3$ different datasets\footnote{Only real valued dimensions are considered and for the `forest fire' dataset dimensions are excluded, where data have more than half $0$ entries.} to test this procedure. As in Fig.~\ref{fig:fig4}, fitting is repeated $5$ times for random permutations if training and test set. For the `Thyroid' dataset, one of the $5$ fits is excluded, because the importance sampling yielded poor approximation ${\rm std}(I)>I\times10^{-2}$. The training sets contain $400$ to $6000$ data points with $5$ to $9$ dimensions. The results for KDE are not reported, since it is always outperformed by the GMM. Fig.~\ref{fig:fig5} demonstrates combining the GMM and VB algorithm results in an improvement of the log test likelihood $\ell_{\rm test}$ compared to using only GMM. Average relative improvements of $\ell_{\rm test}$ are $8.9\ \%$ for `Forest Fire', $4.1\ \%$ for `Thyroid', and $1.1\ \%$ for `Wine' dataset.

\section{DISCUSSION}\label{sec:discussion}
We have shown how inference for a nonparametric, GP based, density model can be made efficient. In the following we would like to discuss various possible extensions but also limitations of our approach.

\paragraph{Choice of base measure} As we have shown for applications to real data, 
the choice of the base measure is quite important, especially for the sampler and for high dimensional problems. While many datasets might favour a normal distribution as base measure, problems with outliers might favour fat tailed densities. In general, any density which can be evaluated on the data space $\X$ and which allows
for efficient sampling, is a valid choice as base measure $\pi(\bx)$ in our inference approach for the GP 
density model. Any powerful density estimator which fulfils this condition could provide 
a base measure which could then potentially be improved  by the GP model. It would e.g. be interesting to apply
this idea to neural networks~\cite{larochelle2011neural,uria2014deep} based estimators. 
Other generalisations of our model could consider alternative data spaces $\X$. One might e.g. think of
specific discrete and structured sets $\X$ for which appropriate Gaussian processes could be defined 
by suitable Mercer kernels. 

\paragraph{Big data \& high dimensionality} Our proposed Gibbs sampler suffers from  
cubic scaling in the number of data points and is found to be already impractical for problems with hundreds of observations. This could potentially be tackled by using sparse (approximate) GP methods for the sampler (see~\cite{samo2015scalable} for a potential approach).  On the other hand, 
the proposed VB algorithm scales only linearly with the training set size and can be applied to problems with several thousands of observations. The integration of stochastic variational inference into our method could potentially increase this limit~\cite{hoffman2013stochastic}.

Potential limitations of the GP density model are given by high dimensional problems. If approached naively, the 
combination of the sparse GP approximation and the numerical integration using importance sampling is expected to yield bad approximations in such cases.\footnote{Potentially in such cases other sparsity methods~\cite{gal2015improving} might be more favourable.} If the data is concentrated on a low dimensional submanifold
of the high--dimensional space, one could still try to combine our method with other density estimators providing a base measure $\pi(\bx)$ that is adapted to this submanifold, to allow for tractable GP inference.

\subsubsection*{Acknowledgements}

CD was supported by the Deutsche Forschungsgemeinschaft (GRK1589/2) and partially funded by Deutsche Forschungsgemeinschaft (DFG) through grant CRC 1294 ``Data Assimilation'', Project (A06) ``Approximative Bayesian inference and model selection for stochastic differential equations (SDEs)''.

\bibliographystyle{unsrt}
\bibliography{bib}

\newpage
\appendix

\section{THE CONDITIONAL POSTERIOR POINT PROCESS}\label{app:poisson proof}
Here we prove  that the conditional posterior point process in Equation~\eqref{eq:latent poisson gibbs} again is a Poisson process using Campbell's theorem~\cite[chap. 3]{kingman1993poisson}. For an arbitrary function $h(\cdot,\cdot)$ we set  
$H \doteq \sum_{(\bx,\omega)\in\Pi}h(\bx,\omega)$. We calculate the characteristic functional
\begin{equation}
\begin{split}
& \EE{\phi_\lambda}{\left. e^{H}\right\vert g,\lambda} = \\
& \frac{\EE{\phi_\lambda}{\left . \prod_{(\omega,\bx)\in \Pi} e^{f(\omega,-g(\bx))
 + h(\bx,\omega)}\right\vert g, \lambda}}{\exp\left(\int_{\X\times \R^+} \left(e^{f(\omega,-g(\bx))}-1\right)
\phi_\lambda(\bx,\omega) d\omega d\bx \right)} = \\
& \frac{\exp\left\{\int_{\X \times \R^+} \left(e^{f(\omega,-g(\bx))+ h(\bx,\omega)} - 1\right) \phi_\lambda(\bx,\omega)  d\omega d\bx \right\}}
{\exp\left(\int_{\X \times \R^+} \left(e^{f(\omega,-g(\bx))}-1\right)
\phi_\lambda(\bx,\omega) d\omega d\bx \right)} = \\
& \exp\left\{\int_{\X \times \R^+} \left(e^{h(\bx,\omega)} - 1\right)e^{f(\omega,-g)}
\phi_\lambda(\bx,\omega) d\omega d\bx \right\} = 
\\
& \exp\left\{\int_{\X \times \R^+} \left(e^{h(\bx, \omega)} - 1\right)\Lambda(\bx,\omega) d\omega d\bx \right\},
\end{split}
\end{equation}
where the last equality follows from the definition of $\phi_\lambda(\bx,\omega)$ and the tilted Polya--Gamma
density. Using the fact that a Poisson process is uniquely characterised by its generating function 
this shows that the conditional posterior  $p(\Pi\vert g, \lambda)$ is a marked Poisson process.

\section{VARIATIONAL LOWER BOUND}\label{app:LB}
The full variational lower bound is given by
\begin{equation}
\begin{split}
\LB(q) = & \sum_{n=1}^N\left\lbrace \EE{Q}{\ln \lambda} + \ln \pi(\bx_n) + \EE{Q}{f(\omega_n,g(\bx_n))} \right. \\
 & \left. - \ln\cosh\left(\frac{c_n}{2}\right) + \frac{c_n^2}{2}\EE{Q}{\omega_n} \right\rbrace \\
& + \int_\X\int_{\R^+} \left\lbrace \EE{Q}{\ln \lambda} + \EE{Q}{f(\omega,-g(\bx))} - \ln \lambda_1 \right. \\
& \left. - \ln \sigma(-c(\bx)) - \ln\cosh\left(\frac{c(\bx)}{2}\right) - \frac{c(\bx)^2}{2}\omega \right. \\
&\left. -\frac{c(\bx) - g_1(\bx)}{2} + 1\right\rbrace \Lambda_1(\bx,\omega)d\omega d\bx \\
& - \EE{Q}{\lambda} + \EE{Q}{\ln\frac{p(\lambda)}{q(\lambda)}} + \EE{Q}{\ln\frac{p(\gs)}{q(\gs)}}.
\end{split}
\end{equation}

\end{document}